\documentclass[times,1p]{elsarticle}
\usepackage{comment}
\usepackage[colorlinks=true,linkcolor=black, citecolor=blue, urlcolor=blue]{hyperref}
\usepackage{url}

\usepackage{siunitx}
\usepackage{float}
\usepackage{subcaption}
\usepackage{tikz}
\usepackage{pgfplots}
\usepackage[english]{babel}
\usepackage{amsmath}
\usepackage{lipsum}  
\usepackage{dsfont}
\usepackage{wrapfig}
\usetikzlibrary{positioning,calc,shapes,arrows,shapes.multipart}
\usepackage{enumitem}

\definecolor{color1bg}{HTML}{1f77b4}
\definecolor{color2bg}{HTML}{ff7f0e}
\definecolor{color3bg}{HTML}{2ca02c}
\definecolor{color4bg}{HTML}{d62728}

\tikzset{
	papDecision/.style = {
		diamond,
		fill = color3bg,
		aspect=2,
		draw, 
		text width = 20 mm, 
		align = center, 
		text badly centered,
		inner sep = 1 pt,
		font=\ttfamily\footnotesize,
		minimum width = 30mm,
		minimum height = 7mm,
	},
	papStart/.style = {
		rectangle,
		fill = color1bg,
		draw, 
		align = center, 
		text width = 3cm, 
		text badly centered,
		inner sep = 4 pt,
		rounded corners=10pt,
		font=\ttfamily\footnotesize,
		minimum width = 30mm,
		minimum height = 7mm,
	},
	papEnd/.style = {
		rectangle,
		fill = color1bg,
		draw, 
		align = center, 
		text width = 3cm, 
		text badly centered,
		inner sep = 4 pt,
		rounded corners=10pt,
		font=\ttfamily\footnotesize,
		minimum width = 30mm,
		minimum height = 7mm,
	},
	papData/.style = {
		trapezium,
		draw, 
		align = center, 
		text width = 20 mm, 
		text badly centered,
		inner sep = 4 pt,
		trapezium left angle=70,
		trapezium right angle=110,
		font=\ttfamily\footnotesize,
		minimum width = 30mm,
		minimum height = 7mm,
	},
	papPredProc/.style = {
		draw,
		rectangle split,
		rectangle split horizontal,
		rectangle split parts = 3,
		rectangle split empty part width=-8pt,
		align = center, 
		text badly centered,
		font=\ttfamily\footnotesize,
		minimum width = 30mm,
		minimum height = 7mm,
	},
	papProcess/.style = {
		rectangle,
		fill = color2bg,
		draw,
		align = center, 
		text width = 3cm, 
		text badly centered,
		font=\ttfamily\footnotesize,
		minimum width = 30mm,
		minimum height = 7mm,
	},
	papLine/.style = {
		draw,
		-stealth,
		font=\ttfamily\footnotesize,
	},
}

\begin{document}
    \title{Inverse Deep Learning Ray Tracing for Heliostat Surface Prediction}
    \author[1,2]{Jan Lewen\corref{cor1}}
    \ead{jan.lewen@dlr.de}
    \author[1]{Max Pargmann}
    \author[2]{Mehdi Cherti}
    \author[2]{Jenia Jitsev}
    \author[1]{Robert Pitz-Paal}
    \author[1]{Daniel Maldonado Quinto}

    \cortext[cor1]{Corresponding author}
    \address[1]{German Aerospace Center (DLR), Institute of Solar Research, Linder Höhe, D-51147 Köln, Germany}
    \address[2]{Research Center Jülich, Jülich Supercomputing Centre, Wilhelm-Johnen-Straße, 52428 Jülich, Germany}
    
    \begin{abstract}
        Concentrating Solar Power (CSP) plants play a crucial role in the global transition towards sustainable energy. A key factor in ensuring the safe and efficient operation of CSP plants is the distribution of concentrated flux density on the receiver. However, the non-ideal flux density generated by individual heliostats can undermine the safety and efficiency of the power plant. The flux density from each heliostat is influenced by its precise surface profile, which includes factors such as canting and mirror errors. Accurately measuring these surface profiles for a large number of heliostats in operation is a formidable challenge. Consequently, control systems often rely on the assumption of ideal surface conditions, which compromises both safety and operational efficiency. \newline In this study, we introduce inverse Deep Learning Ray Tracing (\textit{iDLR}), an innovative method designed to predict heliostat surfaces based solely on target images obtained during heliostat calibration. Our simulation-based investigation demonstrates that sufficient information regarding the heliostat surface is retained in the flux density distribution of a single heliostat, enabling deep learning models to accurately predict the underlying surface with deflectometry-like precision for the majority of heliostats. Additionally, we assess the limitations of this method, particularly in relation to surface accuracy and resultant flux density predictions. Furthermore, we are presenting a new comprehensive heliostat model using Non-Uniform Rational B-Spline (\textit{NURBS}) that has the potential to become the new State of the Art for heliostat surface parameterization. Our findings reveal that iDLR has significant potential to enhance CSP plant operations, potentially increasing the overall efficiency and energy output of the power plants.
    \end{abstract}
    
    
    \maketitle
    \newlist{abbrv}{itemize}{1}
    \setlist[abbrv,1]{label=,labelwidth=1in,align=parleft,itemsep=0.1\baselineskip,leftmargin=!}
    
    \section*{List of Abbreviations}
    \sectionmark{List of Abbreviations}
    
    \begin{abbrv}
    \item[CSP]          Concentrating Solar Power
    \item[STJ]          Solar Tower Jülich
    \item[DLR]          Deutsches Zentrum für Luft- und Raumfahrt
    \item[iDLR]			inverse Deep Learning Ray Tracing
    \item[GAN]          Generative Adversarial Network
    \item[GPU]          Graphics Processing Unit
    \item[NURBS]        Non-Uniform Rational B-Spline 
    \item[MAE]          Mean Absolute Error
    \item[Q1/3]         Quartile 1/3
    \item[IQR]          Interquartile Range
    \item[CAD]          Computer Aided Design
    \item[ACC]          Accuracy
    \item[SOTA]         State of the Art
    \end{abbrv}
    
    \section{Introduction} \noindent
        In order to enhance the economic viability of solar tower power plants, it is crucial to enhance their efficiency. A viable approach for achieving this is through the optimization of the flux density distribution on the receiver. In commercial plants this can be done by optimal aimpoint control. e.g. using the ant-colony optimization meta-heuristic~\cite{Belhomme2013_aimpoints, OBERKIRSCH2023_closed_loop_aimt_point} or deep learning models~\cite{Wu2023} to calculate a flux density for each heliostat using ray tracing and optimize in a next step the flux density distribution on the receiver as a superposition of those. Moreover, using the knowledge about the focal spot shape and its position on the receiver even higher efficiency gains can be achieved. Even simple assumptions can increase the energy output by up to 20\%~\cite{Zhu2022}. However, in commercial power plants aimpoint distribution is mostly static or just to mitigate temperature peaks and gradients that may cause material wear, e.g. by using the Vant Hull algorithm~\cite{vant1996real_a, vant1996real_b}.  This can be explained by the fact that exact information about the heliostats is not easily accessible. Inherent mirror errors and uncertainties about the heliostat's precise geometry model contribute to the individuality of each heliostat's flux density shape and the tracking error. \newline 
        The mirror error is constituted of roughness, slope, and canting errors. The roughness, arises from sub-micrometer flaws on the reflective surface. Slope error measures the deviations of the mirror surface from its ideal shape. Canting error, on the other hand, reflects misalignment among mirror facets. Among these errors, slope and canting are particularly crucial for the flux density~\cite{vant1991concentrator, lovegrove2012concentrating, ARANCIBIABULNES2017673_Review}. A precise knowledge about the heliostats mirror error can be loaded in a ray tracing environment to predict the flux density of the heliostat under certain environmental conditions accurately~\cite{Belhomme2009, Ulmer2011-gp} . \newline 
        The most prominent and state-of-the-art method for obtaining specific surface profiles of heliostats is deflectometric measurement~\cite{Ulmer2011-gp, el2019review}. This method involves capturing camera images of stripe patterns with diverse frequencies projected onto a Lambertian target and subsequently reflected from the heliostat. However, this approach faces obstacles in the heliostat field due to factors such as dew, wind, and that the measurement has to be conducted at night with long exposure times. Alternatively, some methods utilize a laser~\cite{photogrammetry_and_laserscanning, monterreal2017improved}, while others employ photogrammetry~\cite{Shortis1996, Pottler2005-yb, pottler2005photogrammetry, roger2010heliostat, bonanos2019heliostat} or flux mapping \cite{MARTINEZHERNANDEZ2023112162}. However, up for today these measurements are unreliable or cost intensive. For a complete review the reader is referred to~\cite{ARANCIBIABULNES2017673_Review}.
                
        As a result, considerable efforts are made to extract additional information about each heliostat from more readily available data sources. \citet{Zhu2022} have developed a post-installation calibration procedure to find four geometrical parameter per facet by optimization using only target images of the corresponding heliostat. Even this simple helistat model achieved the above mentioned efficency gains. These results underline the potential of including an accurate flux density prediction in the aim point optimization. A key drawback of this method is the simple and specific heliostat model, assuming only canting and focussing error. For example the heliostats at Solar Tower Jülich (STJ), Germany are not focussed and the important surface features that influences the flux density are the waviness of the heliostat and the bending of the facets at the edges and corners of each facets. Those surface features are not predictable by the approach of~\citet{Zhu2022}.
        The latest method for enhancing the precision of flux density predictions is pioneered by Martinez et al~\cite{MARTINEZHERNANDEZ2023112162}. They have devised a technique to reconstruct the surface of heliostats using focal spots during daylight hours. However, this reconstruction capability decreases when the distance between the heliostat and target exceeds 6 meters. To address this limitation, additional measures such as implementing a moving target are necessary.
        \newline \newline 
        ~\citet{Pargmann2023, PARGMANN2023111962} proposed a versatile approach for both heliostat calibration and surface reconstruction, leveraging differentiable ray tracing. This optimization-based method aims to utilize target images for both tasks, thus offering a cost-effective solution that doesn't necessitate additional hardware and can be fully automated. However, it struggles to accurately predict heliostat surfaces due to the inherently underdetermined and ill-posed nature of the problem. \newline
        Consequently, there is a need for a cost-effective, and reliable method capable of predicting heliostat surfaces during regular power plant operations. \newline \newline
        Generative deep learning models have emerged as a prominent focus in computer science over recent years, marked by remarkable breakthroughs facilitated by architectures such as StyleGAN~\cite{karras2019stylebased}, Large Language Models~\cite{sparksAI} and Diffusion Models~\cite{rombach2022highresolution}. Common to generative models is that they gather knowledge about their specific domain during the training and can give predictions even in underdetermined regimes. \newline
        In this work we present a deep learning approach that aims to take target images as input and predicts the corresponding heliostat surface. In contrast to deflectometry, laser scanning, photogrammetry and flux mapping it comes at low cost as no additional hardware or manual work is required for the inference. 
    
    \section{Workflow of inverse Deep Learning Ray Tracing} 
\begin{figure}[h]
            \includegraphics[width=\columnwidth]{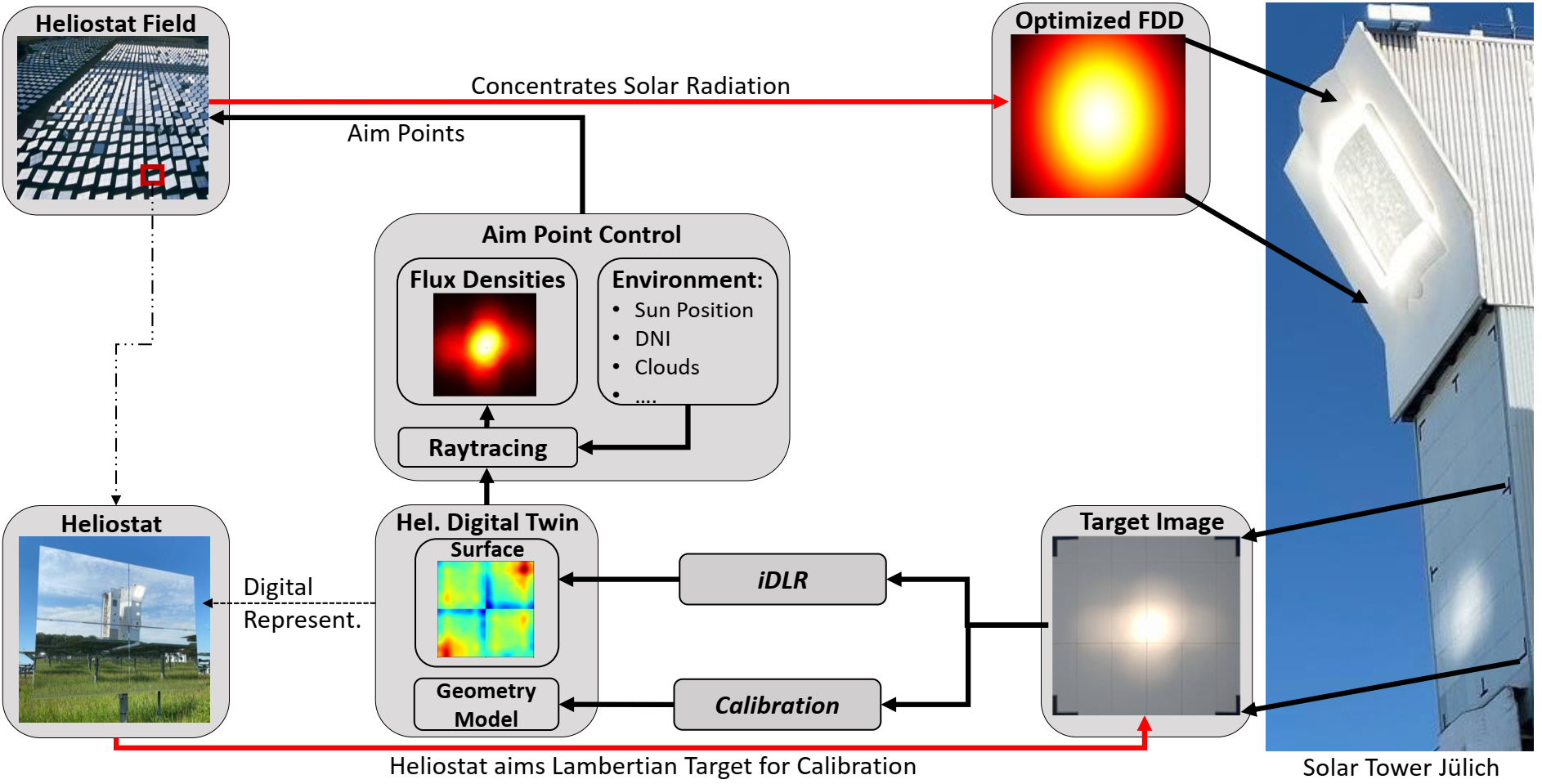}
            \caption{Inverse Deep Learning Ray Tracing (iDLR) embedded in the regular power plant operation. The heliostats are sequentially and fully automatically focused on a Lambertian target and target images are taken for the calibration (Camera-Target Method). Those target images contain information about the heliostat surface. Leveraging a deep learning model, this information is extracted and utilized to predict the heliostat surface, without the necessity of introducing new hardware or executing measurement which are not done yet during regular power plant operation. Consequently, the existing power plant systems and software can now operate seamlessly with an improved heliostat digital twin. This enhancement enables the power plant to operate more efficiently by achieving a more optimal flux density distribution (FDD) on the receiver.}
            \label{fig:method}
        \end{figure}

\noindent
        The primary measurements executed at CSP plants are target images employed for calibration with the \emph{Camera-Target Method}~\cite{stone1986targetmethod}, offering a direct observation of a heliostat's flux density. The flux density shape is caused by the precise heliostat surface characteristics and known parameters such as sun and heliostat positions, and hence the flux density contains information about the surface. However, predicting surfaces from the heliostat’s flux density is highly ill-posed and underdetermined due to facets canting and the overlap of reflected and scattered rays on deformed heliostat surfaces and hence highly challenging. \newline
        This study aims to use deep learning models to predict the heliostat surface using the heliostat’s target images as input. This approach is the inverse direction to physical ray tracing and hence we term the method \textit{inverse Deep Learning Ray Tracing} (\textit{iDLR}). Figure \ref{fig:method} shows the integration of iDLR in the routine power plant operation. Deep learning models have the capability to learn typical surface deformations constrained by material properties and mechanical construction of the mirror surface. This unique ability enables them to predict heliostat surfaces even within the underdetermined regime, leveraging the surface information learned during training. With this capability iDLR should perform better in this underdetermined problem than comparable physical algorithms like differentiable ray tracing~\cite{Pargmann2023}. \newline \newline
        \noindent 
    
        Before training a deep learning model, the first question that arises is which data can be used. A purely real data approach would involve collecting real data pairs of heliostat surfaces and target images. However, there are significant drawbacks to this approach. First, the high costs associated with surface measurement for all existing methods. Second, the method should be applicable to commercial power plants that do not have a surface measurement setup. Finally, the substantial data demand for training generative deep learning models may pose challenges due to the limited number of heliostats in most fields. \newline
        To circumvent these limitations, our methodology aims to train on a semi-artificial dataset. The initial step involves collecting real surface measurements of the heliostat type from any power plant. Subsequently, a model is trained using a dataset comprising augmented real surface data and simulated flux density data obtained through ray tracing. This model is presented in this work. \newline \newline
        In the subsequent phase, the simulated model can be applied to real-world target images using deep learning sim-to-real transfer techniques, to bridge the gap between simulated flux densities and target images \cite{DomainRand_Tobin, DomainRand_Peng2018, DomainRand_Sadeghi_Levine, DomainRand:OpenAI}.

        \subsection{Comprehensive NURBS Heliostat Model}\noindent
            \label{subsec:heliostat_model}
            We employ a heliostat model proposed by \citet{Pargmann2023}, which integrates a traditional geometric model with an innovative spline approach to parameterize the reflecting surface of the heliostat. Current state-of-the-art (SOTA) methods represent heliostat surfaces as a point cloud of normal vectors~\cite{Ulmer2011-gp}, providing precise representation but at the cost of a high number of parameters. For instance, the deflectometry used at the Solar Tower Jülich predicts approximately 80,000 normal vectors per facet (1.6 m $\times$ 1.25 m). This presents a challenge for training deep learning models due to the high computational resource demands and the potential decrease in model precision from the large number of free parameters.
            
            \citet{Pargmann2023} employ a differentiable Non-Uniform Rational B-Spline (NURBS) within a differentiable ray tracing routine (code published at \cite{diffraytracer_code}) as a trainable parameter to fit a heliostat surface to a given flux density. NURBS are the industrial standard to represent complex geometries with few parameters in Computer-Aided Design (CAD) \cite{DevaPrasad2022}. We adapt this method by fitting the NURBS parameter against the normal vectors from the deflectometry measurement through a gradient-based optimization process. The NURBS parameter subsequently serves as a representation of the heliostat surface.
            
            Initially, we reduce the number of NURBS parameters. By using one NURBS per facet, fixing the xy-position of each control point on a predefined grid, and setting the NURBS weight to zero, the heliostat surface is parameterized by only 4x8x8 z-control points. These z-control points are then fitted using the differentiable formulation of \citet{Pargmann2023} and gradient descent against the normal vector cloud of the deflectometry measurement. As a result, the z-control points of the NURBS surface act as a comprehensive representation of the heliostat surface, reducing the number of free parameters from nearly 1 million to just 256—a reduction of 99.97\%. This significantly enhances the efficiency of training machine learning models that involve heliostat surfaces. Furthermore, the NURBS representation is drastically more memory and computationally efficient than the normal point cloud. A single deflectometry measurement's normal vectors occupy 8.6 MB, while the NURBS representation requires only 7 KB, a reduction of 99.91\%.
            
        \subsection{Data Acquisition and Augmentation} \noindent
        \label{subsec:Data}

\begin{wrapfigure}{r}{7cm}
                \includegraphics[width=7cm]{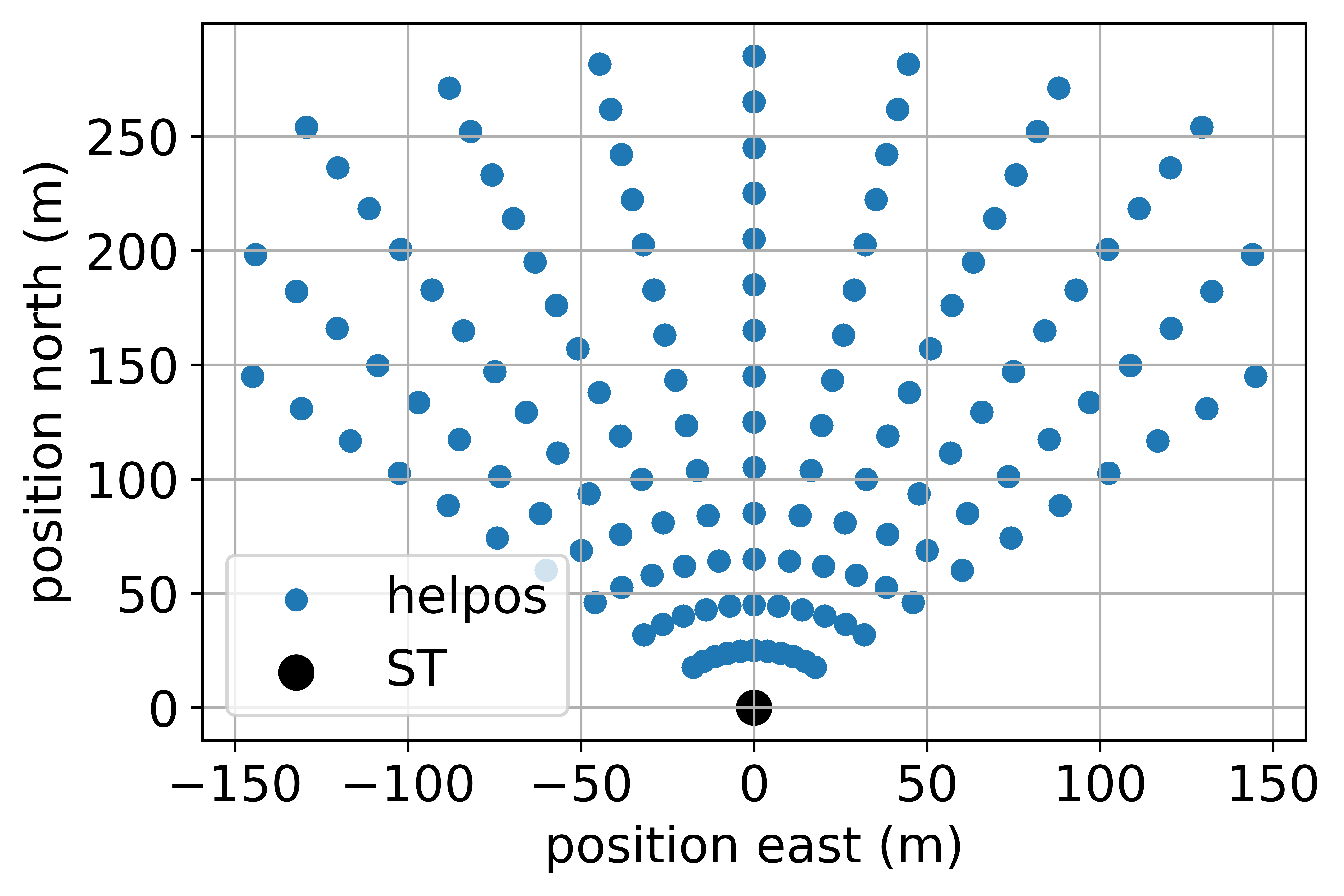}
                \caption{The simulative heliostat field that was used for validating the model. The solar tower (ST) is placed at the origin. At each heliostat position (helpos) the same 32 real measured heliostats were placed.}
                \label{fig:helpos}
            \end{wrapfigure}

            The utilized dataset comprises stripe pattern deflectometry measurements conducted at the STJ, encompassing a total of 458 heliostat surfaces. A split into training data (428), validation (32) and testing data (32) was made, ensuring the exclusion of surfaces from validation and testing heliostats in the model training. For each heliostat, the z-control points of the NURBS were computed as detailed in Section \ref{subsec:heliostat_model} and act as surface representation. As there is a constant unknown offset for the surface of each facet, the mean value of the surface from each facet was set to zero. \newline The dataset comprises an unusually extensive collection of deflectometry data. Despite its size, it remains relatively small for effectively training a neural network. To address this limitation, two types of data augmentation were applied. Firstly, the heliostat surface was rotated by 180~degrees. Secondly, the weighted average between two randomly selected surfaces was calculated using $ z_{augm} = \alpha*z_1 + (1-\alpha)*z_2$ with $\alpha \in (0,1)$. This approach ensures training on a diverse artificial dataset, avoiding the use of surfaces with unrealistic features given the physical constraints of the material and mechanical construction. In total, around 160.000 artificial heliostat surfaces were generated. \newline
            For testing and validation the 32 heliostats were placed in total 4544 times over the field, with a distance of up to 300m to the tower and with a maximum azimuth to the sides of 45~degrees (see Figure~\ref{fig:helpos}). These heliostat positions were chosen as they match those at the STJ.
            \newline
            Subsequently, environmental parameters such as sun position, sun shape, heliostat aim point, and position were randomly drawn in accordance with the geographical position and heliostat field of the STJ. Finally, eight flux densities were simulated for each heliostat surface using the ray tracer developed by~\citet{Pargmann2023}. Those flux densities were centred around the center of mass, cropped onto a size of \SI{4}{m}~x~\SI{4}{m} and finally normalized that the sum of the discrete points of the flux density equals to 100. \newline 
            Finally, the training, validation and test set are formed by the simulated flux densities, the sun position and the heliostat position as input parameter and the z-control points of the NURBS as the surface representation the model will predict. 
        \subsection{Model and Training} 
\begin{figure}[h]
                \includegraphics[width=\columnwidth]{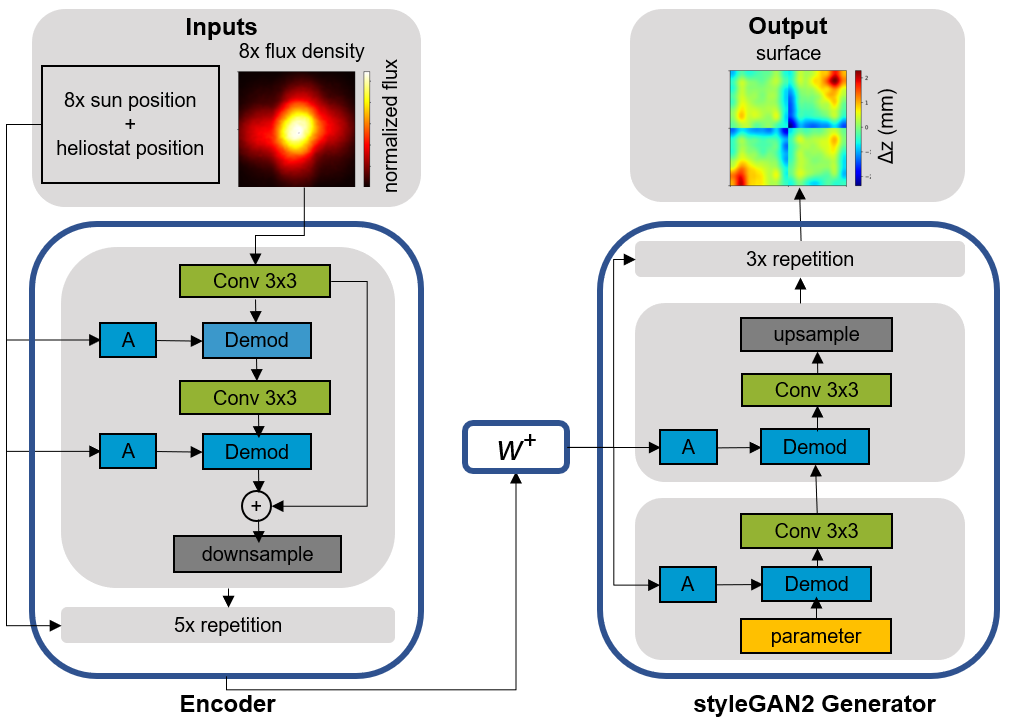}
                \caption{The employed model utilizes up to 8 flux densities, along with corresponding sun and heliostat positions, as inputs. These scalar inputs undergo an affine transformation A, followed by weight demodulation, to map them into the image data stream. Subsequently, convolutional neural layers process those. The resulting \textit{w+} latent space is then fed into the generator of the styleGAN2 architecture, generating the cartesian representation of the surface spline.}
                \label{fig:network}
             \end{figure}

\noindent
            The nature of the problem suggests the adoption of an Encoder-Decoder architecture (see Figure \ref{fig:network}). For the decoder, we employ the StyleGAN2 architecture~\cite{karras2020analyzing} which is highly successful in generating data of a certain domain. The encoder is tasked with fusing two distinct input data streams: the eight flux densities, parameterized as an 8x64x64 tensor, and scalar data containing the corresponding sun positions and the position of the heliostat in the field. 
            \noindent
            The input flux densities undergo processing by a convolutional encoder, while \textit{weight demodulation}, as elucidated in~\citet{karras2020analyzing}, is employed to integrate the scalar information. The number of channels is kept constant by 256 in the encoder. The latent space size is 3*32 using the extended \textit{w+} latent space of the styleGAN2 architecture introduced by~\citet{richardson2021encoding}. The latent surface representation in the latent space is fed in the decoder which generates a cartesian representation of the surface spline. The total number of trainable parameter is 3.5 million. \newline \newline
            The model underwent training utilizing the semi-artificial dataset detailed in Chapter \ref{subsec:Data} for a total of 50 epochs. By using also less than eight input flux densities and sun positions during the training we ensure that the model can predict surfaces for any number of input flux densities up to eight.

    \section{NURBS for Surface Representation: Quality and Computational Efficiency}
            To assess the quality of the NURBS parameterization, we compare it to the deflectometry normal vector point cloud. 
            \begin{figure}[h]
                \centering
                \begin{subfigure}[t]{0.45\textwidth}
                    \includegraphics[width=\textwidth]{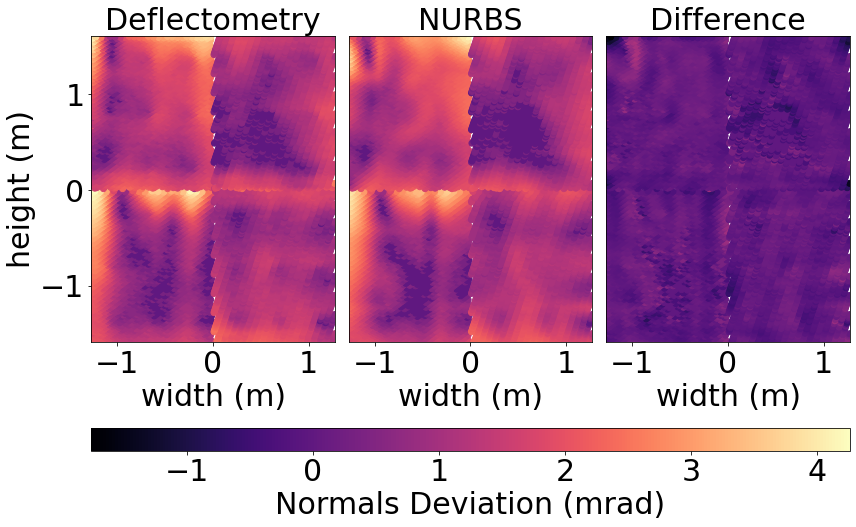}
                    \caption{Left: normal deviation to the ideal flat surface of a deflectometry measurement, middle: normal deviation to the ideal flat surface of the NURBS parameterization of the deflectometry measurement, right: difference plot}
                    \label{fig:normals_nurbs}
                \end{subfigure}
                \hfill
                \begin{subfigure}[t]{0.45\textwidth}
                    \includegraphics[width=\textwidth]{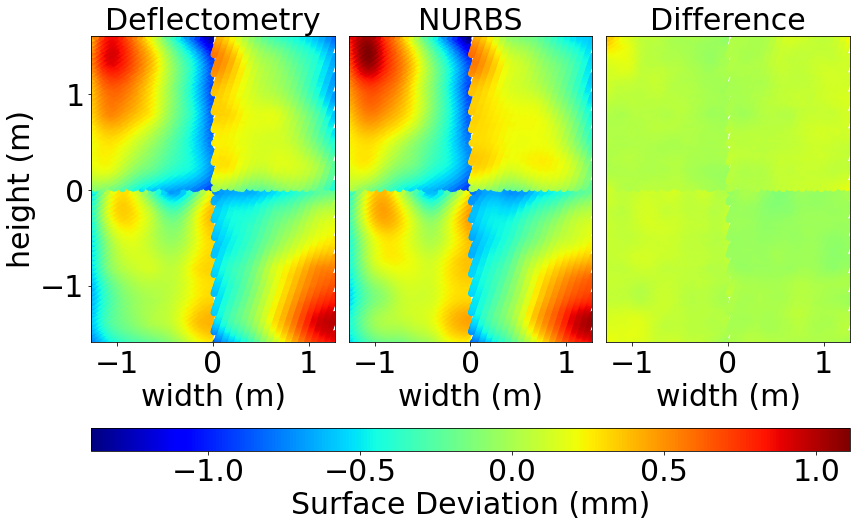}
                    \caption{left: deflectometry surface constructed by a conventional integration-based algorithm from the deflectometry normals, middle: NURBS based heliostat surface, right: difference plot}
                    \label{fig:surface_nurbs}
                \end{subfigure}
                \caption{Illustration of the NURBs fitting procedure for an individual heliostat. The left panel depicts the normal vectors resulting from the NURBS parameterization, while the right panel displays cartesian representations of the surface.}
            \label{fig:nurbs}
            \end{figure}
            Figure \ref{fig:normals_nurbs} compares the normal vectors from deflectometry measurements with those from the NURBS surface, demonstrating minimal differences and underscoring the accurate parameterization of the heliostats' normal vectors using the 4x8x8 NURBS z-control points. In Figure \ref{fig:surface_nurbs}, the surface deviation of the heliostat from the ideal assumption is depicted, with a conventional integration-based algorithm estimation on the left and the NURBS parameterization on the right. The strong alignment between both underscores the robust parameterization achieved with the NURBS parameters. To quantify the quality of the NURBS parameterization, we calculate the angular deviation between the normal vectors from the NURBS spline and the deflectometry measurements. 81\% of the NURBS normal vectors have an angular deviation of zero mrad from the original deflectometry normal vectors within machine accuracy. The remaining 19\% form an extremely skewed tail, with the largest outlier being 10.2~mrad. While the majority of the normal vectors are fitted perfectly, the outliers have mainly three causes: the NURBS normals must be smoother than the original, sharp edges in the contour lines are rounded, and at the boundary of the facets, the NURBS parameterization becomes slightly less accurate due to boundary effects.
        
\begin{figure}[h]
                \centering
                \begin{subfigure}[t]{0.49\textwidth}
                    \includegraphics[width=\textwidth]{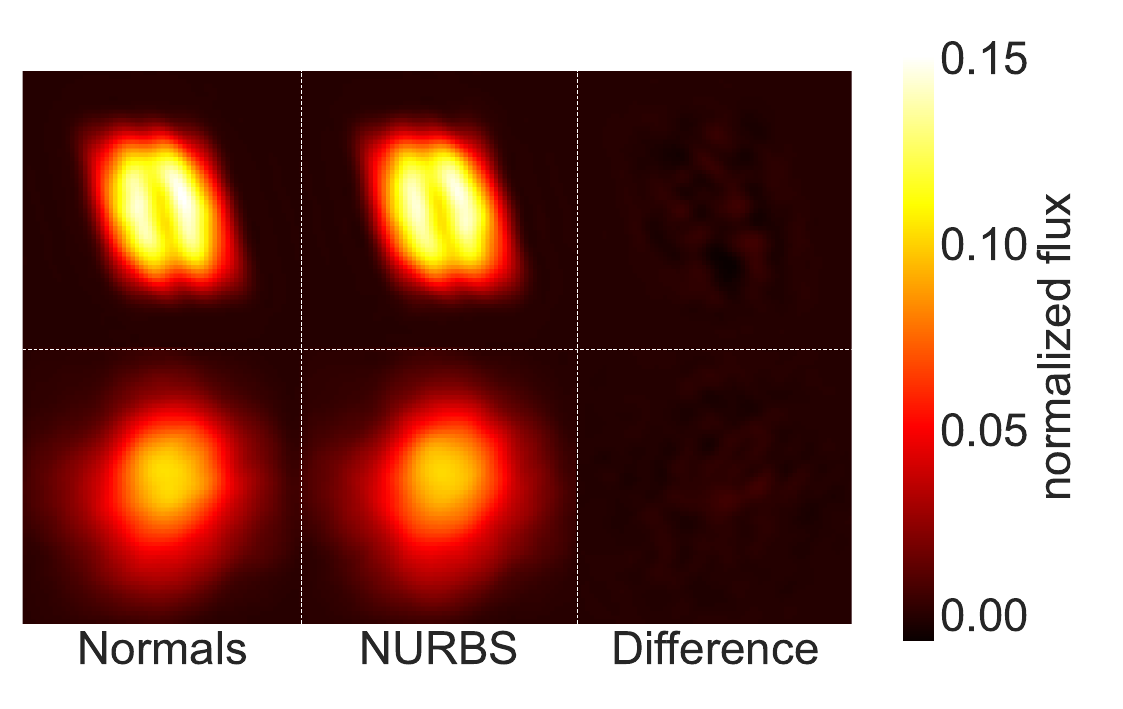}
                    \caption{Simulated flux densities for the heliostat with the smallest outlying fitted normal vector (upper row) and the largest outlying fitted normal vector (lower row) are shown. The left column displays the flux density prediction using the normal vector cloud from the deflectometry measurement, while the middle column presents the flux density resulting from the NURBS parameterization. The right column illustrates the difference between these two flux density predictions.}
                    \label{fig:flux_NURBS}
                \end{subfigure}
                \hfill
                \begin{subfigure}[t]{0.49\textwidth}
                    \includegraphics[width=\textwidth]{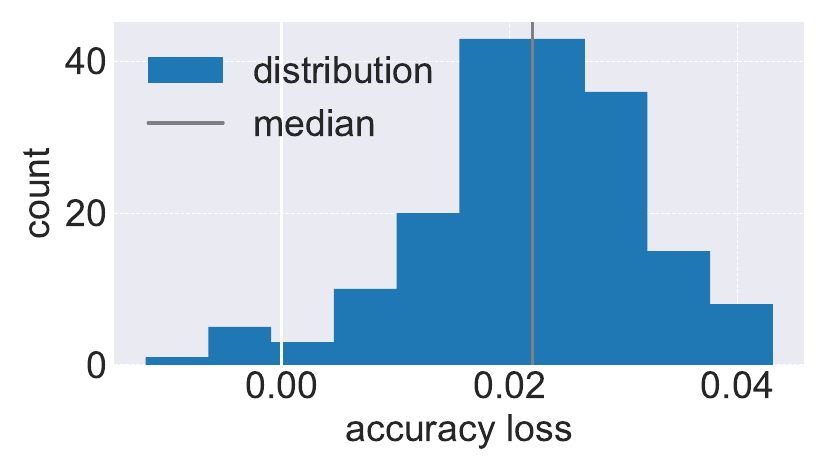}
                    \caption{The accuracy loss in flux density prediction that can be attributed to the NURBS parameterization. The distribution is centred around the median accuracy loss of 2.2\%. Hence there is a significant, but relatively small loss due to the NURBS parameterization.}
                    \label{fig:hist_nurbs}
                \end{subfigure}
            \label{fig:NURBS_analysis}
            \end{figure}

            Next, we compare the flux density predictions based on the normal vector cloud and the NURBS spline loaded into the ray tracer proposed by \citet{Pargmann2023}. 
            Figure \ref{fig:flux_NURBS} shows the flux density prediction from deflectometry normals and NURBS for the heliostat with the largest outlying normal vector deviation (10.2~mrad) and the smallest (2~mrad). In both cases, the flux densities match strongly with nearly no visible difference, highlighting the robust surface parameterization of the NURBS. To quantify the quality of the flux density prediction, the following accuracy metric is calculated:
            \begin{equation}
                \text{{ACC}}_\text{GT,pred} = \frac{{\sum |\phi_{\text{{GT}}} - \phi_{\text{{pred}}}|}}{{\sum |\phi_{\text{{GT}}}|}}
            \end{equation}
            As ray tracing involves Monte-Carlo Sampling, slight variations in flux densities are expected when ray tracing the same scene twice. To account for this, the flux density for the normal vector cloud was calculated twice $\phi_{\text{{Normals1/2}}}$, and the loss of accuracy caused by the NURBS parameterization is calculated as:
            \begin{equation}
                \text{{loss}} = 1 - (\text{ACC}_\text{NURBS, Normals1} + (1 - \text{ACC}_\text{Normals1, Normals2}))
            \end{equation}
            Figure \ref{fig:hist_nurbs} illustrates the accuracy loss in flux density predictions attributable to the NURBS parameterization for 184 deflectometry measurements conducted at STJ. These 184 measurements constitute the subset with more than 99\% of their mirror surface measured successfully (see Section~\ref{subsec:Data} for more details on data). The distribution centers around its mean $\text{loss}_\text{NURBS} = 2.2$, indicating a significant but relatively small loss due to the NURBS.
        \newline \newline
        The substantial parameter reduction by 99.97\% (from more than one million to 256) is essential for the implementation of iDLR. Although it is technically feasible to train generative models with over a million pixels, the computational expense is prohibitive; for example, the styleGAN architecture requires 41 days and 4 hours of training on a Tesla V100 GPU for 1024x1024 pixels \cite{karras2019stylebased}. Additionally, training such models is highly challenging due to the large number of trainable parameters, which complicates regularization.
        
        This parameter reduction significantly enhances memory efficiency, decreasing the memory requirement by 99.91\% from 8.6 MB to 7 KB per heliostat. This improvement is critical for two main reasons. First, deep learning models often necessitate hundreds of thousands of training samples, making a memory-efficient data representation imperative. Second, large concentrated solar power (CSP) plants may comprise up to 200,000 heliostats, necessitating an efficient surface representation for real-time flux density predictions during operation, which includes heliostat digital twins with surface information.
        
        The deviations between the NURBS normals and the original normals have resulted in a minor flux density accuracy loss of 2.2\%. This loss is particularly minimal when contrasted with the accuracy of the ideal heliostat assumption at the STJ, which is $\text{ACC}_\text{ideal}=67\%$. Consequently, the potential gain in flux density prediction accuracy substantially outweighs the loss introduced by the NURBS parameterization. It is feasible to further reduce the mrad and flux density loss by using additional NURBS control points. However, the accuracy achieved with our current parameterization is considered sufficient for our purposes.
        
        The proposed NURBS parameterization represents a significant advancement over the current SOTA in heliostat surface parameterization, specifically the normal vector cloud. Key advantages include enhanced memory and computational efficiency, which are crucial for large power plants employing a heliostat digital twin that incorporates surface information for real-time flux density prediction. Moreover, the NURBS method allows for a reduced number of free parameters in machine learning models and provides a differentiable formulation. Consequently, all training losses and validation metrics in this work are based on the NURBS parameterization. Overall, our findings suggest that NURBS will become the new SOTA for heliostat surface parameterization.

    \section{Results}
    \subsection{Surface Prediction}
    \begin{figure}[h]
                \includegraphics[width=\columnwidth]{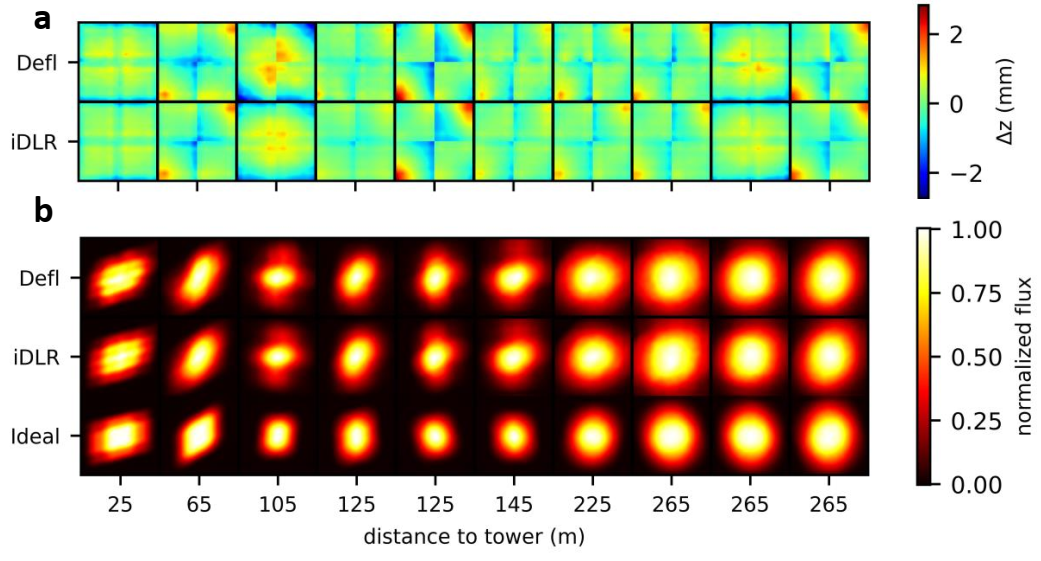}
                \caption{\textbf(a): The presented surface predictions are randomly selected. The top row illustrates the ground truth obtained through deflectometry, while the bottom row depicts the predictions generated by iDLR. \textbf{b}: The provided flux density predictions correspond to the heliostats shown above. The top row presents the ground truth, derived through ray tracing deflectometry surfaces, while the middle row showcases predictions generated by ray tracing the iDLR predictions. As a point of reference, the bottom row displays flux density predictions derived from the ideal heliostat assumption. It is noteworthy that the flux density prediction for poor iDLR surface predictions (see third heliostat from the left) is just as good as for those with a very good surface prediction. This phenomenon can be attributed to the inherent underdetermination of the problem. The flux density predictions generated from iDLR surface predictions significantly surpass those based on the ideal heliostat assumption.}
                \label{fig:predictions}
             \end{figure}
\begin{figure}[h]
        \begin{subfigure}[t]{0.45\textwidth} 
            \includegraphics[width=\textwidth]{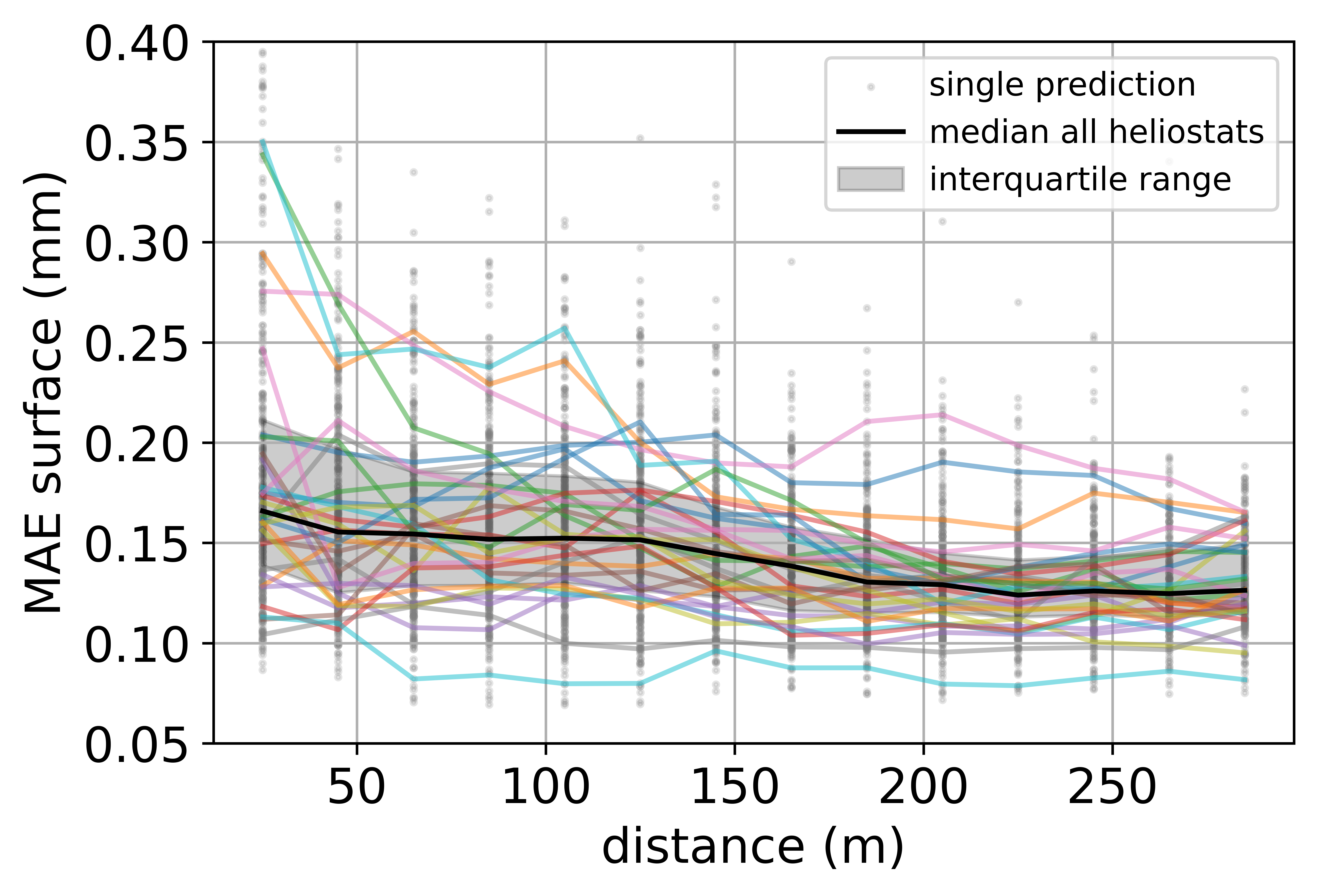}
            \caption{The presented scatter plot illustrates the Mean Absolute Error per pixel (MAE) of surface predictions as a function of distance. Each data point represents the MAE for an individual prediction. The colored lines correspond to the MAE for a specific heliostat positioned at various locations across the field, while the black lines depict the median MAE from all heliostats.}
            \label{fig:MAE_over_distance}
        \end{subfigure} \hfill
        \begin{subfigure}[t]{0.45\textwidth}
            \includegraphics[width=\textwidth]{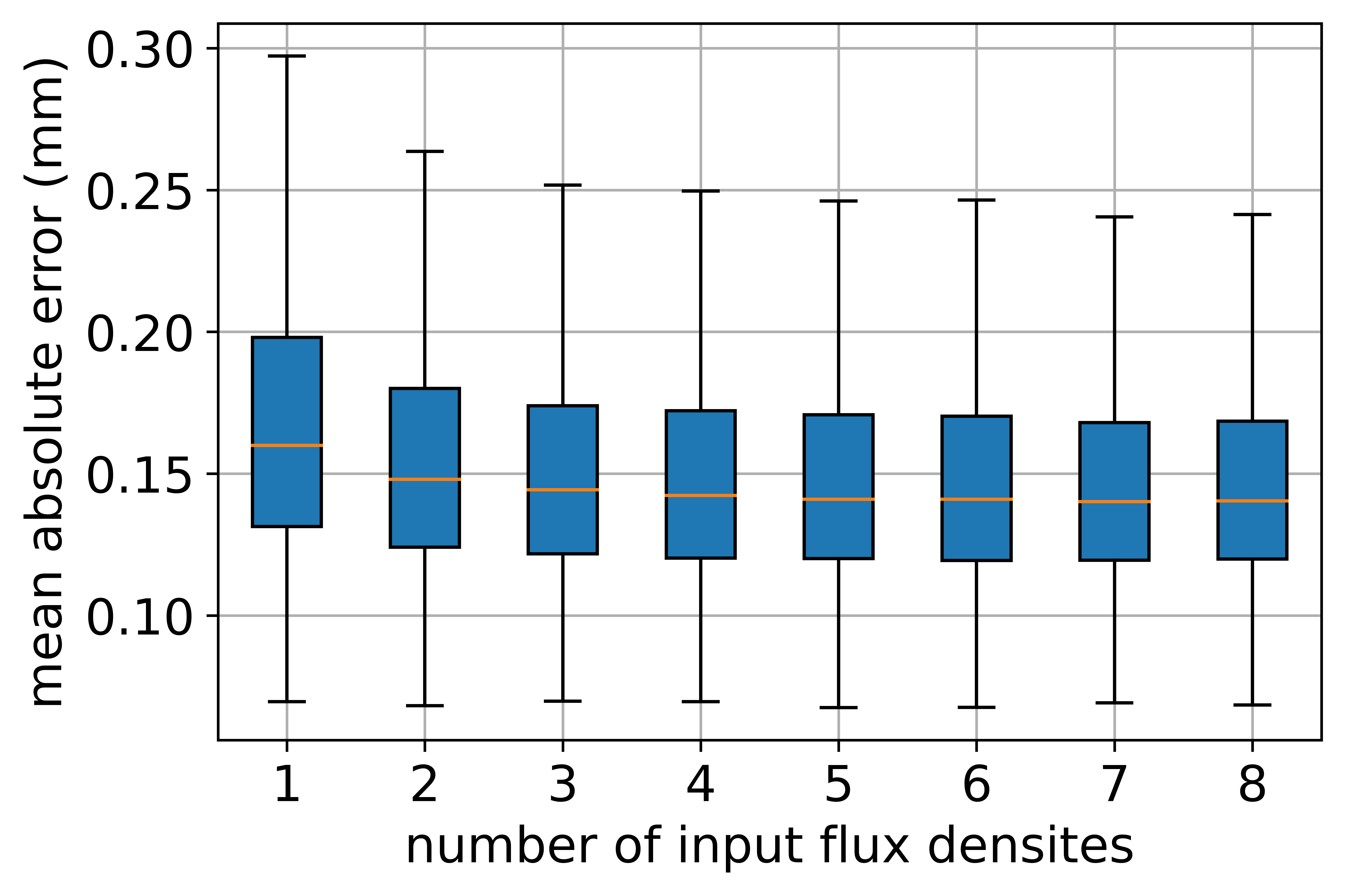}
            \caption{The impact of the number of input flux densities on surface prediction is shown as a function of the MAE of number of input flux densities. The model exhibits substantial improvement when adding more flux densities and goes then into a saturation.}
            \label{fig:sunpos}
        \end{subfigure}
    \end{figure}

\noindent
        In Figure \ref{fig:predictions}a, ten randomly selected surface predictions by iDLR are shown, whereby seven simulated flux densities were used as an input. The mean absolute error per control-point (MAE) between the predicted surface and the deflectometry-derived surface has a median of 0.14~mm with (Mininmum/Quantile~1/Quantile~3/Maximum) of (0.07/0.12/0.17/0.7)~mm. For comparison the typical surface deviation falls within the range of -2 mm to 2 mm, but exhibiting considerable variability among heliostats. As illustrated in Figure \ref{fig:predictions}a, a strong alignment is evident between the predicted surface and the deflectometry ground truth surface. Notably, the model demonstrates accurate prediction of heliostat waviness, representing surface features with the highest spatial frequency that is representable with the NURBS spline. Figure \ref{fig:MAE_over_distance} shows the performance of the model in dependency to the heliostats distance to the tower (a few outliers are not displayed). The surface predictions become significantly better at larger distances up to the considered maximum distance of 300m. However, no analogous effect is identified concerning the azimuth of the heliostat position. Figure \ref{fig:sunpos} illustrates the model's surface reconstruction performance in relation to the number of input flux densities. Notably, a substantial enhancement in precision is observed with an increased number of input flux densities. However, given that the model was trained with a maximum capacity for processing eight flux densities, the potential impact of utilizing an even larger number of input flux densities on the results remains unclear as the saturation of the curve could be at a smaller MAE.
        
     \noindent

    \subsection{Flux density prediction} 
\begin{figure}
        \includegraphics[width=\columnwidth]{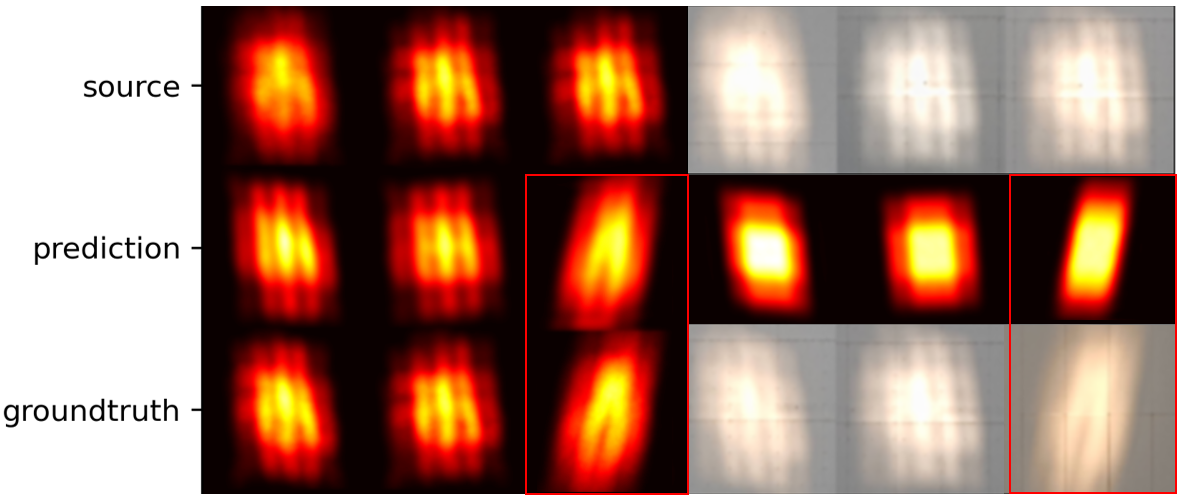}
        \caption{Comparison of simulated flux densities arising from ray tracing deflectometry and iDLR surfaces with target images obtained at the Solar Tower Jülich. The top row displays three simulated flux densities (left) corresponding to three target images (right). These flux densities were used to predict the heliostat surfaces (not shown). Subsequently, flux density predictions for three new sun positions were generated with a ray tracer (middle row, left). The three flux densities on the right are generated under the ideal heliostat assumption. The bottom row presents the ground truth flux densities along with their corresponding target images. Notably, the red-marked target images were captured from an additional target positioned 18m west and 15m higher than the source target. It is visible that iDLR and ray tracing can successfully predict the flux density in this spatial extrapolation scenario, a crucial capability given that the receiver is located in a different plane than the target.}
        \label{fig:flux_target}
    \end{figure}

\noindent
    In the subsequent phase, the predicted surfaces serve as a basis for forecasting flux densities of heliostats using ray tracing. Randomly selected flux density predictions are visualized in Figure \ref{fig:predictions}b and compared to the ideal heliostat assumption. A high match between the flux densities evoking from iDLR and deflectometry surfaces is visible across all shown heliostats. Remarkably, this does not only hold for the heliostats with accurate surface prediction (compare Figure \ref{fig:predictions}a) but also for those with a poor surface reconstruction (the third heliostat from the left). This phenomenon can be attributed to the inherent underdetermination of the problem, as surface deformations that significantly change the facets mean normal vector (especially canting errors) have a stronger effect on the flux density shape for heliostats at larger distances. To quantify the quality of the flux density predictions, the mean accuracy is computed as described in section~\ref{subsec:heliostat_model}. The median accuracy of the flux density prediction of the model is 0.92 with (Min/Q1/Q3/Max) of (0.43/0.90/0.94/0.98), proving a very accurate and reliable flux density prediction using the surfaces predicted with the model for most predictions. Generally, even fine details can be predicted using the method, however some small deviations from especially inhomogeneous and exceptional flux densities are possible. For comparison, the ideal heliostat assumption of a flat surface without deformations achieves a median accuracy of only 0.67 with (Min/Q1/Q3/Max) of (0.37/0.59/0.74/0.9). This is not only on median 0.25 (-0.08/0.18/0.33/0.55) less accurate then the prediction with iDLR and ray tracing, but also the interquartile range (IQR) of the ideal heliostat prediction $(\text{IQR}_{\text{ideal}} = 0.15)$ is more than three times as high as those from iDLR ($\text{IQR}_{\text{iDLR}} = 0.04$) and make iDLR to the more precise and reliable model. Over the whole test data set consisting of 32 simulated heliostats on 142 positions, the ideal heliostat flux density was more accurate than the iDLR prediction for only four heliostats, showing a failure rate of iDLR based flux density prediction around 1\textperthousand~in the presented simulated environment. \newline \newline 
    Figure \ref{fig:flux_target} compares simulated flux densities from ray tracing deflectometry measurements and iDLR predictions with the corresponding target image taken at the STJ.
     \noindent
    The heliostat is positioned at 4.4m west and 25m north of the tower. Three source flux densities are used to predict the heliostats surface and three new flux densities are predicted. The simulated flux densities correspond to target images that were taken for calibration purpose at the STJ (right side). The high match between ray tracing deflectometric measured heliostats and the target images taken during the calibration measurement shows that the information necessary to predict the surfaces are in the target images as well as in the ray tracing simulation, emphasizing the possibility to transfer the simulative model to real data by machine learning techniques. \newline
    Moreover, Figure \ref{fig:flux_target} illustrates a spatial extrapolation flux density prediction and its comparison to a real target image. The STJ incorporates a secondary target positioned 18m west and 15m higher than the main target. The high correspondence between the flux density prediction for spatial extrapolation and the ground truth, including the real target image, underscores the capability of the predicted surface to not only forecast flux density on the source target but also on another target plane. This is important as receivers are situated in a different plane than the source targets and hence the flux density prediction on the receiver is a spatial extrapolation.
    \section{Discussion} \noindent
    The results obtained through simulation underscore the significant ability of iDLR. The method demonstrates its capability to predict heliostat surfaces with high accuracy using simulated flux densities, despite the inherently undetermined and ill-posed nature of the inverse problem. While the majority of the reconstructed surfaces closely align with the deflectometry ground truth (median MAE of 0.14~mm with (Min/Q1/Q3/Max) of (0.07/0.12/0.17/0.7)~mm), it is important to acknowledge a potential drawback: there exists a risk of less accurate surface predictions compared to deflectometry measurements for a minority of heliostats. This is particularly evident for heliostats with a small amount of input flux densities and those in close proximity to the tower, where surface predictions may exhibit deviations from deflectometry ground truth. This discrepancy arises due to the influence of canting errors and surface deformation, leading to changes in the mean normal vector of facets. The impact of these factors on flux density is stronger for heliostats located further away from the tower, providing more information about those surface deviations. However, this relationship does not hold for surface errors with higher frequency heliostat surface features, such as waviness. With higher distances to the tower, the features in the flux density caused by waviness diminishes and the flux density becomes more smooth. Notably, the model demonstrates the ability to predict higher frequency heliostat features even for heliostats at greater distances, indicating the utilization of learned static material constraints to make accurate predictions rather than relying solely on conserved information in the flux density. \newline \newline 
    In contrast to the singular existing alternative for heliostat surface prediction from target images, the differentiable ray tracer proposed by ~\citet{Pargmann2023}, our findings demonstrate substantial improvement. While the differentiable ray tracer can only predict the surface for one singular heliostat with pronounced, artificially introduced surface deformations, iDLR is able to predict a wide range of measured, realistic heliostat surfaces. The problem of the differentiable ray tracer lies in the underdetermined nature of the problem as it does not acquire any preknowledge about the physical constraints of the surface deformations. Our approach successfully navigates this challenge by learning the physical constraints of the heliostat surfaces, thereby narrowing the solution space and enabling precise predictions despite the inherent underdetermination and ill-posed characteristics of the problem. Furthermore, iDLR predictions come at real time while the predictions of differentiable ray tracing is depended on the number of epochs and number of input flux densities, resulting in computation times in the range of a few minutes up to some hours per heliostat. \newline 
    On the other side, we identified three drawbacks of iDLR. Firstly, while the model yields real-time predictions, its training process is notably time-intensive. The execution of ray tracing and subsequent flux density processing consumes approximately 0.1 seconds per iteration. Consequently, the generation of the training dataset demanded approximately 45 hours of computation on a single GPU (NVIDIA A100). Subsequent training and testing of the model on the same GPU required additional 12 hours. Despite this initial investment in time, the eventual computational costs are anticipated to be outweighed by iDLR's real-time inference capabilities, rendering it superior to optimization-based algorithms like differentiable ray tracing over already short periods of usage. Secondly, to achieve accurate surface predictions, the model necessitates training on realistic heliostat surfaces, which can be obtained through measurement or simulation. While surface measurement incurs associated costs, it's noteworthy that each power plant intending to implement iDLR need not possess a measurement setup; rather, only one setup per heliostat type is requisite at any power plant or even at the manufacturer side. Alternatively, circumventing the necessity for heliostat surface measurements, finite-element simulation of heliostats could be employed to predict potential deformations induced by construction-induced stress. Simple surface simulation, that do not consider physical factors like adding gaussian deformations to the ideal heliostat, were tried out and it was found that the trained model has problems to predict real heliostat surfaces from the simulated flux densities, as it does not learn the heliostat surface space with its physical constraints. Lastly, the model may encounter challenges in accurately predicting particularly rare or unique surface deformations that substantially deviate from those encountered during its training phase. Such out of distribution predictions pose a notable hurdle, necessitating ongoing refinement and adaptation of the iDLR framework. \newline \newline 
    The minimum requisite quantity of measured heliostat surfaces for model training is contingent upon two primary factors. Firstly, it relies on the diversity present in surface deformations, as the iDLR model necessitates comprehending a broader surface space, thereby requiring a larger dataset for effective learning. Secondly, it is influenced by the generalizability of a distinct set of deflectometry measurements. Biases often manifest within datasets, such as an overrepresentation of deflectometry measurements from heliostats situated in close proximity to the tower, owing to the ease of application in these areas. Given the likely existence of correlations between mirror error and heliostat position (e.g. larger canting errors occurring at closer positions), a biased dataset may diminish the predictive accuracy of the model, even when training set sizes remain constant.
    \newline \newline
    When comparing to existing methods for surface measurements such as deflectometry~\cite{Ulmer2011-gp}, laser scanning, photogrammetry~\cite{Shortis1996, photogrammetry_and_laserscanning, pottler2005photogrammetry} or flux mapping \cite{MARTINEZHERNANDEZ2023112162}, iDLR stands out as a software-only approach for surface prediction. It relies on hardware already employed in commercial power plants and eliminates the need for active human involvement during execution, making it considerably more cost-effective. However, as mentioned above, this is traded against the risk of having a less accurate surface prediction for a minority of the heliostats. \newline \newline 
    The flux density predictions obtained through ray tracing the iDLR prediction significantly outperforms the ideal heliostat assumption (accuracy of iDLR is 0.92 compared to accuracy of ideal heliostat assumption of 0.67). This highlights the potential benefits of replacing the ideal heliostat model with the iDLR prediction in the operation of a solar power plant as aim-point optimization strategies reliant on ray tracing a heliostat model, as seen in works like~\cite{ OBERKIRSCH2023_closed_loop_aimt_point, Wu2023, Zhu2022,Belhomme2013_aimpoints_ant}, can experience enhancements when the ideal heliostat assumption is substituted with the iDLR surface predictions. \newline \newline
    To successfully apply the trained model to real-world data, the limited availability of surface data for training poses significant challenges. Relying solely on real data for training is impractical due to the requirement for surface measurements of a large number of heliostats in the corresponding power plant. Consequently, the preferred approach involves transferring the simulated model presented here to real data using deep learning techniques like domain randomization ~\cite{DomainRand_Tobin, DomainRand_Peng2018, DomainRand:OpenAI, DomainRand_Sadeghi_Levine}. Key latent parameters, such as varying sun shapes~\cite{Neumann1999, Buie2003}, soiling~\cite{Berg1978, Roth1980}, mirror surface roughness~\cite{Bonanos2012}, heliostat geometry models, and the disparities between simulated flux density and target images due to non-ideal Lambertian properties and background radiation reflection, can be randomized during simulation. This can ensure the robustness of the simulated model against differences between simulation and real-world data. \newline 
    An alternative strategy for transferring the model to real-world data involves incorporating differentiable ray tracing formulations or utilizing a deep learning ray tracer within the workflow. This approach would enables close-loop training, using target images exclusively after pretraining on simulated data.
    \section{Conclusion}\noindent
    \label{sec:conclusion_outlook}
    We have introduced inverse Deep Learning Ray Tracing (iDLR), an innovative and cost-effective methodology for predicting heliostat surfaces from target images, employing deep learning to solve the inverse direction to ray tracing. The simulation results indicate that iDLR holds considerable promise as an in-situ approach for predicting heliostat surfaces from target images. The predicted surfaces demonstrate competitive accuracy compared to deflectometry surfaces for the majority of heliostats, all while incurring minimal costs. Integrating the predicted surface into ray tracing aim-point optimization strategies has the potential to enhance the flux density distribution on the receiver, thereby optimizing the overall efficiency of power plant operations. 

    \section*{Competing interests}\noindent
    The authors declare no competing interests.

    \section*{Declaration of generative AI in scientific writing}\noindent
    During the preparation of this work the author(s) used large language models (LLM) (GPT3.5/4 of openAI and the LLMs by deepL) in order to improve the language. After using this tool/service, the author(s) reviewed and edited the content as needed and take(s) full responsibility for the content of the publication.
    \section*{Acknowledgements}\noindent
    We gratefully acknowledge the use of data from the Solar Tower Jülich, a research power plant operated by the German Aerospace Center (DLR). Additionally, we extend our thanks for the funding provided by the Helmholtz Association (HGF) for the GANCSTR project (Grant Number ZT-I-PF-5-069), under which this research was conducted. 
    \bibliography{lit}
    \bibliographystyle{elsarticle-num-names}
    \appendix
    \newpage
    
\end{document}